\documentclass[11pt]{article}

\usepackage[T1]{fontenc}
\usepackage{lmodern}
\usepackage{microtype}
\usepackage{geometry}
\usepackage{authblk}

\usepackage{amsmath,amssymb,amsfonts}
\usepackage{mathtools}
\usepackage{bm}

\usepackage{amsthm}
\theoremstyle{plain}
\newtheorem{theorem}{Theorem}[section]
\newtheorem{lemma}[theorem]{Lemma}
\newtheorem{proposition}[theorem]{Proposition}

\theoremstyle{definition}

\newtheorem{assumption}{Assumption}[section]
\theoremstyle{remark}
\newtheorem{remark}{Remark}[section]

\numberwithin{equation}{section}

\usepackage{float}
\usepackage{xurl}

\usepackage{graphicx}
\usepackage{subcaption}
\usepackage{booktabs}
\usepackage{array}
\usepackage{tabularx}
\newcolumntype{Y}{>{\raggedright\arraybackslash}X}

\usepackage[inline,shortlabels]{enumitem}
\setlist{nosep}
\setlist[itemize]{leftmargin=1.2em}
\setlist[enumerate]{leftmargin=1.2em}

\usepackage{algorithm}
\usepackage{algpseudocode}

\usepackage{tikz}
\usetikzlibrary{arrows.meta, positioning, matrix, fit, calc, backgrounds, shadows, shapes.geometric}

\usepackage[numbers,sort&compress]{natbib}

\usepackage[colorlinks=true,linkcolor=magenta,citecolor=blue,urlcolor=blue]{hyperref}
\usepackage[nameinlink,capitalize,noabbrev]{cleveref}

\newcommand{\R}{\mathbb{R}}

\newcommand{\E}{\mathbb{E}}
\newcommand{\Var}{\mathrm{Var}}
\newcommand{\Cov}{\mathrm{Cov}}

\newcommand{\eg}{e.g.}

\newcommand{\train}{\mathcal{D}}

\newcommand{\vzero}{\bm{0}}

\newcommand{\va}{\bm{a}}

\newcommand{\vd}{\bm{d}}
\newcommand{\ve}{\bm{e}}

\newcommand{\vg}{\bm{g}}

\newcommand{\vj}{\bm{j}}

\newcommand{\vm}{\bm{m}}

\newcommand{\vs}{\bm{s}}

\newcommand{\vu}{\bm{u}}
\newcommand{\vv}{\bm{v}}
\newcommand{\vw}{\bm{w}}
\newcommand{\vx}{\bm{x}}

\newcommand{\vz}{\bm{z}}

\newcommand{\mA}{\bm{A}}

\newcommand{\mC}{\bm{C}}

\newcommand{\mH}{\bm{H}}
\newcommand{\mI}{\bm{I}}
\newcommand{\mJ}{\bm{J}}
\newcommand{\mK}{\bm{K}}

\newcommand{\mM}{\bm{M}}

\newcommand{\mQ}{\bm{Q}}
\newcommand{\mR}{\bm{R}}

\newcommand{\mZ}{\bm{Z}}

\DeclareMathOperator*{\argmin}{arg\,min}

\DeclareMathOperator{\sign}{sign}
\DeclareMathOperator{\diag}{diag}

\newcommand{\sigmoid}{\sigma}

\newcommand{\norm}[1]{\left\lVert #1 \right\rVert}
\newcommand{\abs}[1]{\left\lvert #1 \right\rvert}

\newcommand{\Ls}{\mathcal{L}}  

\newcommand{\cF}{\mathcal{F}}      

\title{
Incremental Gauss-Newton Descent for Machine Learning
}

\author[1]{Mikalai Korbit}
\author[1]{Mario Zanon}

\affil[1]{IMT School for Advanced Studies Lucca\par Lucca, Italy}

\date{}

\begin{document}

\maketitle

\begin{abstract}

Stochastic gradient updates are widely used for their efficiency and scalability, 
but their effective step sizes can depend strongly 
on feature scaling and local model sensitivity.
Gauss-Newton methods address such scale effects through curvature information, 
but in their standard mini-batch form they require matrix-vector products, 
linear solves, or structured approximations. 
This paper studies the special case of scalar-output losses 
evaluated one sample at a time.
In this setting, the generalized Gauss-Newton matrix has rank at most one, 
and its only possible nonzero curvature direction is aligned with the
stochastic gradient.
As a result, the damped Gauss-Newton direction reduces 
to a closed-form scalar normalization of the sample gradient. 
The resulting update, Incremental Gauss-Newton Descent (IGND), 
requires no curvature matrix storage, factorization, or iterative linear solve. 
We derive the update, characterize its behavior, 
and relate it to normalized gradient descent, 
adaptive first-order methods, stochastic Polyak step sizes, 
and mini-batch Gauss-Newton updates. 
Under explicit smoothness, alignment, and stochastic approximation assumptions,
we prove a stationarity result for the IGND update. 
Experiments on supervised learning, 
a controlled test of scale robustness, 
and a linear-quadratic control case study 
show that IGND improves robustness to sensitivity scaling 
and can be competitive with, or complementary to, 
common stochastic optimizers 
while retaining a simple incremental update.

\end{abstract}

\section{Introduction}\label{sec:intro}

Many learning algorithms are built from incremental stochastic updates: 
a single sample or a small batch is processed, 
a stochastic direction is formed,
and the parameters are updated by taking a step along that direction. 
Stochastic Gradient Descent (SGD) 
is the canonical example~\cite{robbins1951stochastic}. 
Its appeal is clear: 
the update is simple, memory-light, and easy to implement in modern
automatic differentiation frameworks. 
Its weakness is equally familiar: the same learning rate 
can behave very differently after changing feature units,
output scales, or local model sensitivities.

This scale sensitivity is already visible in the simplest linear case. 
At equivalent parameterizations that give the same current prediction, 
if a single active linear feature is multiplied by a factor $\phi$, 
then an SGD parameter step gives 
a first-order prediction change proportional to $\phi^2$.
The optimizer behavior therefore depends not only on the loss 
but also on arbitrary representation choices. 
Adaptive first-order methods such as 
AdaGrad~\cite{duchi2011adaptive}, 
RMSProp~\cite{tieleman2012lecture}, 
AdaDelta~\cite{zeiler2012adadelta},
and
Adam~\cite{kingma2014adam} 
reduce some of this tuning burden by using accumulated gradient statistics.
Their rescaling, however, is history-based and coordinatewise; 
it does not explicitly target the current sample's local output sensitivity, 
which is what controls the induced change in the scalar prediction.

Gauss-Newton and generalized Gauss-Newton (GGN) methods 
are attractive precisely because 
they incorporate local model sensitivity and curvature 
information~\cite{nocedal2006numerical,schraudolph2002fast}. 
In their standard mini-batch or vector-output form, 
these methods involve a curvature matrix and require 
linear solves, matrix-vector products, sketches, factorizations, 
or other approximations.
This makes them expensive for large models.

The key observation of this paper is the following. 
Let $f_t=f_{\vw_t}(\vx_t)$ be a scalar model output at iteration $t$, 
let $\vj_t=\nabla_{\vw} f_{\vw_t}(\vx_t)$ be the output sensitivity vector, 
and let $\ell_{f,t}$ and $\ell_{ff,t}$ be the first and second derivatives 
of the sample loss with respect to the scalar output. 
Then the sample gradient and the corresponding GGN matrix are
\begin{align*}
    \vg_t=\ell_{f,t}\vj_t,
    \qquad
    \mH_t^{\mathrm{GGN}}=\ell_{ff,t}\vj_t\vj_t^\top.
\end{align*}
Both objects are tied to the same output sensitivity vector $\vj_t$: 
the sample gradient lies in $\mathrm{span}\{\vj_t\}$, 
and the GGN matrix has curvature only in the span of $\vj_t$.
Consequently, the sample GGN matrix has rank at most one, 
and the damped Gauss-Newton direction can be written in closed form 
as a scalar rescaling of the stochastic gradient. 
We call the resulting update Incremental Gauss-Newton Descent (IGND). 
Here, ``incremental'' is used in the stochastic optimization 
sense of processing one sample at a time.

The exact derivation applies to scalar-output losses evaluated one sample at a time,
including squared error regression, binary cross-entropy with logits, 
and scalar Bellman error updates when the target 
is treated as fixed during the update.
The rank-one property is a single-sample property: for a mini-batch of 
$b$ scalar-output samples, 
the corresponding GGN matrix generally has rank at most $b$, rather than one.
Batched variants based on averaging directions from individual samples are 
therefore practical extensions, 
not exact mini-batch Gauss-Newton solves.

The main contributions are as follows.
\begin{itemize}
    \item We show that the resulting rank-one Gauss-Newton system admits 
    a closed-form damped solution, 
    yielding IGND as a scalar normalization 
    of the stochastic gradient direction.

    \item We analyze how the IGND update changes 
    the current model output under a local linear approximation
    and relate it to SGD, 
    normalized gradient descent,
    normalized least-mean-squares (NLMS), 
    stochastic Polyak step sizes (SPS),
    diagonal adaptive methods, 
    and mini-batch Gauss-Newton.

    \item We prove a stationarity guarantee for damped IGND under
    smoothness, alignment, moment, and stochastic approximation assumptions.

    \item We evaluate IGND on supervised learning tasks, 
    a controlled test of scale robustness, 
    and a linear-quadratic control case study, 
    showing improved robustness to feature scaling 
    and competitive performance relative to standard stochastic optimizers.

\end{itemize}

\section{Related Work}\label{sec:rw}

\paragraph{Adaptive first-order methods}
SGD and its accelerated variants remain the default optimization methods for
large-scale machine learning. 
Momentum methods use running averages of past
gradients to reduce oscillations and accelerate progress in shallow curvature
directions~\cite{polyak1964some,nesterov1983method,sutskever2013importance}.
Adaptive methods such as AdaGrad, RMSProp, AdaDelta, and Adam use diagonal
rescalings derived from gradient 
statistics~\cite{duchi2011adaptive,tieleman2012lecture,zeiler2012adadelta,kingma2014adam}.
These methods are strong practical baselines, 
and our experiments include both SGD and Adam. 
By contrast, IGND uses a scalar normalization derived from the GGN matrix 
of the current sample,
rather than from accumulated gradient statistics.

\paragraph{Approximate Gauss-Newton and second-order methods}
Exact second-order methods are usually impractical for neural networks 
since forming, storing, and inverting Hessian matrices is too expensive. 
This has led to Hessian-free and approximate-curvature methods, including
Krylov subspace methods, KFAC-style factorizations, and block-diagonal
Gauss-Newton approximations
\cite{martens2010deep,vinyals2012krylov,martens2015optimizing,botev2017practical}.
Classical Levenberg-Marquardt damping modifies Gauss-Newton systems to
improve numerical robustness~\cite{levenberg1944method,marquardt1963algorithm,nocedal2006numerical}.
Stochastic Gauss-Newton and Levenberg-Marquardt variants typically use
mini-batch generalized Gauss-Newton matrices together with conjugate gradient,
low-rank approximations, variance-reduced estimators, or linear-algebra
transformations
\cite{gargiani2020promise,ren2019efficient,tran2020stochastic,brust2021nonlinear,hong2020stochastic,korbit2025exact}.
Natural-gradient methods precondition gradients using Fisher-information
geometry, which is closely related to GGN geometry for common probabilistic
losses~\cite{amari1998natural,martens2020new}. 
Their implementation still requires forming, approximating, or applying
a non-scalar curvature preconditioner.

The structure studied here appears for one scalar-output sample. 
In that case, the generalized Gauss-Newton matrix has rank at most one,
and the damped direction reduces to a closed-form scalar normalization of the
stochastic gradient. 
For a mini-batch of $b$ scalar-output samples, 
the generalized Gauss-Newton matrix has the form $\mJ^\top\mC\mJ$ 
and generally has rank at most $b$, not one. 
Mini-batch stochastic Gauss-Newton and natural gradient methods solve or
approximate this higher-rank system; 
IGND computes the exact solution of the
damped rank-one scalar-output GGN system for each sample.

\paragraph{Normalized methods, NLMS, and stochastic Polyak steps}
Gradient clipping and normalization are common ways to reduce sensitivity to
large gradients~\cite{zhang2019gradient,qian2021understanding}. 
Normalized gradient descent has also been studied as an optimization method in its own
right~\cite{murray2019revisiting}. 
In signal
processing, normalized least-mean-squares (NLMS) rescales the linear
least-squares update by the input norm~\cite{widrow1985adaptive}. 
For a linear model with squared loss, 
undamped IGND reduces to the same basic NLMS normalization.

IGND is also related to stochastic Polyak step sizes (SPS), 
which choose the
step length using the current loss value and a lower bound on the attainable
loss~\cite{loizou2021stochastic,orvieto2022dynamics}. 
For squared loss, zero
lower bound, and inactive clipping/capping constants, 
SPS and IGND coincide up
to conventional constants. 
For binary cross-entropy and other non-quadratic scalar losses, 
both methods remain scalar rescalings of the stochastic gradient direction. 
The step lengths differ: 
IGND uses output curvature $\ell_{ff,t}$ and output
sensitivity $\norm{\vj_t}^2$, 
whereas SPS uses a loss value ratio.

\section{Incremental Gauss-Newton Descent}\label{sec:ignd}

This section derives IGND
from the rank-one generalized Gauss-Newton matrix obtained
for one scalar-output sample.
We first define the sample gradient and GGN matrix, 
then derive the Moore-Penrose
and damped closed-form directions.

\subsection{Problem setting}\label{sec:problem-setting}

Scalars are plain, vectors are bold lowercase, and matrices are bold uppercase. 
For $\va\in\R^n$, $\diag(\va)$ denotes the diagonal matrix with diagonal $\va$. 
Gradients of scalar-valued functions are column vectors. 
We write $\norm{\cdot}$ for the Euclidean norm, 
$\mI$ for the identity matrix, 
and $\vzero$ for the zero vector.

We consider empirical risk minimization for a scalar-valued model
$f_{\vw}:\R^n\to\R$ with parameters $\vw\in\R^d$, 
\eg, a neural network with one output.
Given samples $(\vx_i,y_i)\in\R^n\times\R$ and a loss $\ell(y,f)$,
the objective is
\begin{align}\label{eq:problem}
    \min_{\vw\in\R^d} \; \Ls(\vw),
    \qquad
    \Ls(\vw)
    := \frac{1}{N}\sum_{i=1}^N \ell\big(y_i,f_{\vw}(\vx_i)\big),
\end{align}
where the per-sample loss $\ell(y,f)$ is twice continuously differentiable and convex in the scalar prediction $f$. 
At iteration $t$, one sample $(\vx_t,y_t)$ is selected and the update has the form
\begin{align}\label{eq:generic-update}
    \vw_{t+1}=\vw_t+\alpha_t\vd_t,
\end{align}
where $\alpha_t>0$ is the learning rate and $\vd_t$ is the search direction.

Throughout the derivation, define
\begin{align}\label{eq:ft-jt-def}
    f_t := f_{\vw_t}(\vx_t),
    \qquad
    \vj_t := \nabla_{\vw} f_{\vw_t}(\vx_t),
\end{align}
where $\vj_t$ is the local output sensitivity vector.
We also define
\begin{align}\label{eq:loss-derivatives}
    \ell_{f,t}
    := \frac{\partial \ell}{\partial f}(y_t,f_t),
    \qquad
    \ell_{ff,t}
    := \frac{\partial^2 \ell}{\partial f^2}(y_t,f_t).
\end{align}
Convexity in the scalar output gives $\ell_{ff,t}\ge0$.

\subsection{Rank-one GGN structure}\label{sec:rank-one-ggn}

By the chain rule, the sample gradient is
\begin{align}\label{eq:sample-gradient}
    \vg_t
    := \nabla_{\vw}\ell(y_t,f_{\vw_t}(\vx_t))
    = \ell_{f,t}\vj_t.
\end{align}
The exact sample Hessian decomposes as
\begin{align}\label{eq:hessian-decomposition}
    \nabla_{\vw}^2\ell(y_t,f_{\vw_t}(\vx_t))
    = \ell_{ff,t}\vj_t\vj_t^\top
    + \ell_{f,t}\nabla_{\vw}^2 f_{\vw_t}(\vx_t).
\end{align}
The scalar-output GGN approximation 
keeps the positive semidefinite term $\ell_{ff,t}\vj_t\vj_t^\top$ 
and drops the model-curvature term $\ell_{f,t}\nabla^2_{\vw} f_{\vw_t}(\vx_t)$~\cite{schraudolph2002fast,martens2020new}:
\begin{align}\label{eq:H-ggn}
    \mH_t^{\mathrm{GGN}}
    := \ell_{ff,t}\vj_t\vj_t^\top.
\end{align}
The outer-product form in Eq.~\eqref{eq:H-ggn} 
makes the rank-one structure immediate.
The sample gradient is a scalar multiple of $\vj_t$, so
$\vg_t\in\mathrm{span}\{\vj_t\}$.
The GGN matrix acts only on the same span: for any direction $\vu$ such that
$\vj_t^\top\vu=0$, which entails
\[
    \mH_t^{\mathrm{GGN}}\vu
    =
    \ell_{ff,t}\vj_t(\vj_t^\top\vu)
    =
    \vzero.
\]
Moreover,
\[
    \mH_t^{\mathrm{GGN}}\vj_t
    =
    \ell_{ff,t}\norm{\vj_t}^2\vj_t.
\]
Thus $\mH_t^{\mathrm{GGN}}$ has rank at most one.
It is rank one when $\ell_{ff,t}>0$ and $\norm{\vj_t}>0$;
otherwise it is the zero matrix.
This rank condition is independent of whether the sample gradient vanishes:
if $\ell_{f,t}=0$ while $\ell_{ff,t}>0$ and $\norm{\vj_t}>0$, 
then $\vg_t=\vzero$ but $\mH_t^{\mathrm{GGN}}$ remains rank one.
When $\ell_{f,t}\ne0$, 
the stochastic gradient and the only possible nonzero
GGN curvature direction are both parallel to $\vj_t$.
When $\ell_{f,t}=0$, the current sample gradient vanishes; 
the Moore-Penrose and damped Gauss-Newton directions derived below are then zero.

\subsection{Moore-Penrose and the damped rank-one direction}\label{sec:mp-damped-direction}

The corresponding Gauss-Newton linear system is
\begin{align}\label{eq:formal-gn-system}
    \mH_t^{\mathrm{GGN}} \vd_t=-\vg_t.
\end{align}
Since $\mH_t^{\mathrm{GGN}}$ is singular whenever $d>1$, 
this equation does not define a unique direction. 
In the nondegenerate rank-one case, however, 
the system in Eq.~\eqref{eq:formal-gn-system} 
has solutions since both sides lie in the span of $\vj_t$. 
We choose the Moore-Penrose solution, i.e., 
the minimum-norm solution of the singular linear system~\cite{penrose1955generalized,ben2003generalized,bjorck2024numerical}.
Among all directions consistent with the Gauss-Newton equation, 
this selects the one with the smallest norm in parameter space 
by removing arbitrary components in the null space.
Intuitively, 
directions in the null space do not affect the linearized scalar output, 
so the Moore-Penrose solution keeps only the movement needed along 
the active curvature direction.

\begin{proposition}[Rank-one scalar-output GGN direction]
\label{prop:rank-one-ggn-direction}
Consider the loss for one scalar-output sample,
$\ell_t(\vw)=\ell(y_t,f_{\vw}(\vx_t))$, with quantities defined in
Equations~\eqref{eq:ft-jt-def}--\eqref{eq:H-ggn}.
If $\ell_{ff,t}>0$ and $\norm{\vj_t}>0$, then the minimum-norm GGN direction is
\begin{align}\label{eq:mp-ggn-direction}
    \vd_t^{\mathrm{GGN}}
    :=
    -\big(\mH_t^{\mathrm{GGN}}\big)^{\dagger} \vg_t
    =
    -
    \frac{\ell_{f,t}}
    {\ell_{ff,t}\norm{\vj_t}^2}
    \vj_t.
\end{align}
\end{proposition}
\begin{proof}
Since $\ell_{ff,t}>0$ and $\norm{\vj_t}>0$,
\begin{align}\label{eq:rank-one-pinv}
    \big(\ell_{ff,t}\vj_t\vj_t^\top\big)^{\dagger}
    =
    \frac{1}{\ell_{ff,t}}\frac{\vj_t\vj_t^\top}{\norm{\vj_t}^4}.
\end{align}
Using $\vg_t=\ell_{f,t}\vj_t$ gives
\begin{align*}
    -\big(\mH_t^{\mathrm{GGN}}\big)^{\dagger}\vg_t
    =
    -\frac{1}{\ell_{ff,t}}\frac{\vj_t\vj_t^\top}{\norm{\vj_t}^4}\ell_{f,t}\vj_t
    =
    -\frac{\ell_{f,t}}{\ell_{ff,t}\norm{\vj_t}^2}\vj_t.
\end{align*}
\end{proof}

For geometric intuition, 
the undamped direction in Eq.~\eqref{eq:mp-ggn-direction}
can also be characterized as the minimum-norm solution 
of the rank-one quadratic model; see~\ref{apx:qp-rank-one}.

The Moore-Penrose direction is useful conceptually, 
but the undamped expression is undefined 
whenever the output curvature or sensitivity vanishes. 
We therefore use a damped rank-one system. 
For fixed $\epsilon_c\ge0$ and $\epsilon_{\mathrm{LM}}>0$, 
define the damped direction as the solution of
\begin{align}\label{eq:damped-rank-one-system}
    \left((\ell_{ff,t}+\epsilon_c)\vj_t\vj_t^\top
    +\epsilon_{\mathrm{LM}}\mI\right)\vd_t
    =
    -\vg_t.
\end{align}
The two damping constants regularize different parts of the rank-one system.
The term $\epsilon_{\mathrm{LM}}\mI$ is the usual Levenberg-Marquardt damping: 
it makes the matrix in Eq.~\eqref{eq:damped-rank-one-system} positive definite, 
so the damped direction is uniquely defined even in case the rank-one term vanishes.
The scalar $\epsilon_c$ is added to the output curvature $\ell_{ff,t}$ 
before forming the outer product. 
It is not needed for invertibility when
$\epsilon_{\mathrm{LM}}>0$, but, as the closed form below shows, 
choosing $\epsilon_c>0$ keeps a $\norm{\vj_t}^2$ term in the denominator 
when $\ell_{ff,t}$ is close to zero.

\begin{proposition}[Closed-form damped rank-one direction]\label{prop:damped-direction}
For $\epsilon_c\ge0$ and $\epsilon_{\mathrm{LM}}>0$, the unique solution of Eq.~\eqref{eq:damped-rank-one-system} is
\begin{align}\label{eq:damped-direction}
    \vd_t^{\mathrm{DGN}}
    =
    -
    \frac{\ell_{f,t}}
    {(\ell_{ff,t}+\epsilon_c)\norm{\vj_t}^2+\epsilon_{\mathrm{LM}}}
    \vj_t.
\end{align}
\end{proposition}
\begin{proof}
Let
\[
    \mA_t
    :=
    (\ell_{ff,t}+\epsilon_c)\vj_t\vj_t^\top
    +\epsilon_{\mathrm{LM}}\mI
\]
and define
\[
    D_t
    :=
    (\ell_{ff,t}+\epsilon_c)\norm{\vj_t}^2+\epsilon_{\mathrm{LM}}.
\]
Since $\ell_{ff,t}\ge0$, $\epsilon_c\ge0$, and $\epsilon_{\mathrm{LM}}>0$,
the matrix $\mA_t$ is positive definite, so the solution is unique.
Now set
\[
    \vd_t^\star
    :=
    -
    \frac{\ell_{f,t}}{D_t}\vj_t .
\]
Then
\begin{align*}
    \mA_t\vd_t^\star
    &=
    -
    \frac{\ell_{f,t}}{D_t}
    \left(
    (\ell_{ff,t}+\epsilon_c)\vj_t\vj_t^\top\vj_t
    +\epsilon_{\mathrm{LM}}\vj_t
    \right) \\
    &=
    -
    \frac{\ell_{f,t}}{D_t}
    \left(
    (\ell_{ff,t}+\epsilon_c)\norm{\vj_t}^2
    +\epsilon_{\mathrm{LM}}
    \right)\vj_t 
    =
    -\ell_{f,t}\vj_t
    =
    -\vg_t .
\end{align*}
Thus $\vd_t^\star$ solves Eq.~\eqref{eq:damped-rank-one-system}. 
By uniqueness, it is the desired damped direction.
\end{proof}

\paragraph{Degenerate cases}
The damped update handles the singular cases explicitly. 
If $\vj_t=\vzero$, then $\vg_t=\vzero$ and the update is zero. 
If $\ell_{f,t}=0$, the sample is stationary in output space and the update is zero. 
If $\ell_{ff,t}=0$ but $\ell_{f,t}\ne0$, 
the undamped GGN matrix provides no curvature along the sample direction. 
The damped update becomes
\[
-\frac{\ell_{f,t}}{\epsilon_c\norm{\vj_t}^2+\epsilon_{\mathrm{LM}}}\vj_t .
\]
The $\epsilon_c\norm{\vj_t}^2$ term gives scaling that depends on sensitivity 
when $\epsilon_c>0$, 
while $\epsilon_{\mathrm{LM}}$ prevents singularity.
Losses with $\ell_{ff,t}<0$ are outside the GGN interpretation used here, 
which assumes convexity of the loss in the scalar output.

\subsection{The IGND update}\label{sec:ignd-update}

\begin{algorithm}[t]
\caption{Incremental Gauss-Newton Descent (IGND)}
\label{alg:ignd}
\begin{algorithmic}[1]
\Require Dataset $\train$, initial weights $\vw_0$, learning rate sequence $\{\alpha_t\}$, damping parameters $\epsilon_c\ge0$ and $\epsilon_{\mathrm{LM}}>0$
\State Set $t\gets0$
\Repeat
    \State Sample $(\vx_t,y_t)$ from $\train$
    \State Compute $f_t=f_{\vw_t}(\vx_t)$ and $\vj_t=\nabla_{\vw}f_{\vw_t}(\vx_t)$
    \State Compute $\ell_{f,t}$ and $\ell_{ff,t}$
    \State Compute $\xi_t$ from Eq.~\eqref{eq:xi-reg}
    \State Update $\vw_{t+1}=\vw_t-\alpha_t\xi_t\ell_{f,t}\vj_t$
    \State Set $t\gets t+1$
\Until{a stopping criterion is met}
\end{algorithmic}
\end{algorithm}

Given
\begin{align}\label{eq:xi-reg}
    \xi_t
    =
    \frac{1}
    {(\ell_{ff,t}+\epsilon_c)\norm{\vj_t}^2+\epsilon_{\mathrm{LM}}},
\end{align}
the damped IGND update is
\begin{align}\label{eq:ignd-update}
    \vw_{t+1}
    =
    \vw_t
    -
    \alpha_t\xi_t\ell_{f,t}\vj_t
    =
    \vw_t
    -
    \alpha_t
    \frac{\ell_{f,t}}
    {(\ell_{ff,t}+\epsilon_c)\norm{\vj_t}^2+\epsilon_{\mathrm{LM}}}
    \vj_t.
\end{align}
Equivalently, 
since $\vg_t=\ell_{f,t}\vj_t$, 
the update is a scalar normalization of the stochastic gradient:
\begin{align}\label{eq:ignd-gradient-form}
    \vw_{t+1}=\vw_t-\alpha_t\xi_t\vg_t.
\end{align}
When $\ell_{f,t}\ne0$, 
an implementation can compute the ordinary sample gradient and recover
$\norm{\vj_t}^2=\norm{\vg_t}^2/\ell_{f,t}^2$. 
However, this reconstruction becomes ill-conditioned when $\ell_{f,t}$ is
small. 
Computing $\vj_t$ directly separates the loss derivative $\ell_{f,t}$ from
the model sensitivity $\vj_t$, avoids division by $\ell_{f,t}^2$, 
and leaves small sensitivities to be handled 
by the damping term $\epsilon_{\mathrm{LM}}$.
Algorithm~\ref{alg:ignd} summarizes the update.

\section{Analysis of the IGND Update}\label{sec:analysis}

This section analyzes the scalar normalization induced by the rank-one GGN structure.
We first introduce the first-order prediction change $h_t$ 
and use it to study the scale behavior of IGND.
The same quantity then provides a common basis for comparing IGND 
with normalized gradient descent, NLMS, SPS, diagonal adaptive methods, 
and mini-batch Gauss-Newton methods.
We then specialize the update to common scalar-output losses, 
discuss its computational cost, 
and state the empirical risk stationarity guarantee.

\subsection{First-order prediction change and scale robustness}\label{sec:scale}

Parameter steps are not directly comparable by their Euclidean size alone:
for a scalar-output loss, what matters locally is how much the step changes
the current scalar prediction.
Around the current parameters $\vw_t$, a displacement $\Delta\vw_t$ gives
\[
    f_{\vw_t+\Delta\vw_t}(\vx_t)
    \approx
    f_t+\vj_t^\top\Delta\vw_t .
\]
We therefore define
\begin{align}\label{eq:local-prediction-change}
    h_t
    :=
    \vj_t^\top\Delta\vw_t.
\end{align}
The scalar $h_t$ is the first-order prediction change 
induced by the parameter update.
This is the relevant quantity for scale analysis: 
at equivalent parameterizations, feature scaling can leave the current prediction 
and loss derivatives unchanged while changing the map
from a parameter step to a prediction change.

The sample loss can then be locally approximated by 
a one-dimensional quadratic in $h_t$:
\begin{align}\label{eq:output-quadratic}
    \ell(y_t,f_t+h_t)
    \approx
    \ell(y_t,f_t)+\ell_{f,t}h_t+\frac{1}{2}\ell_{ff,t}h_t^2.
\end{align}
When $\ell_{ff,t}>0$, 
the minimizer of this quadratic model is
\begin{align}\label{eq:output-newton-step}
    h_t^\star=-\frac{\ell_{f,t}}{\ell_{ff,t}}.
\end{align}

For undamped IGND,
\begin{align*}
    h_t^{\mathrm{IGND}}
    &=
    \vj_t^\top
    \left(
    -\alpha_t
    \frac{\ell_{f,t}}{\ell_{ff,t}\norm{\vj_t}^2}
    \vj_t
    \right)
    =
    -\alpha_t\frac{\ell_{f,t}}{\ell_{ff,t}}.
\end{align*}
Thus, when $\alpha_t=1$, undamped IGND realizes the Newton step
$h_t^\star$ in the scalar prediction coordinate.
Among parameter displacements that realize this first-order prediction change,
the Moore-Penrose construction selects the minimum-norm displacement along
$\vj_t$.
With damping,
\begin{align}\label{eq:damped-prediction-step}
    h_t^{\mathrm{IGND}}
    =
    -\alpha_t\ell_{f,t}
    \frac{\norm{\vj_t}^2}
    {(\ell_{ff,t}+\epsilon_c)\norm{\vj_t}^2+\epsilon_{\mathrm{LM}}}
    =
    -\alpha_t\ell_{f,t}
    \frac{1}
    {(\ell_{ff,t}+\epsilon_c)+\epsilon_{\mathrm{LM}}/\norm{\vj_t}^2},
\end{align}
when $\vj_t\ne\vzero$. If $\ell_{ff,t}+\epsilon_c>0$, then
for large sensitivities,
\[
    h_t^{\mathrm{IGND}}
    \to
    -\alpha_t\frac{\ell_{f,t}}{\ell_{ff,t}+\epsilon_c}.
\]
When $\epsilon_c=0$ and $\ell_{ff,t}>0$, 
this is the scalar Newton step scaled by $\alpha_t$.
For very small $\norm{\vj_t}$, 
the Levenberg-Marquardt term dominates the denominator and suppresses the update.

The scale effect can be seen in the simplest one-hot linear case.
Consider $f_{\vw}(\vx)=\vw^\top\vx$ with active feature $\vx=\ve_j$.
Under a coordinatewise rescaling of the representation, 
the active feature becomes
\[
    \tilde{\vx}=\phi\ve_j,
    \qquad
    \phi\ne0,
\]
where $\phi$ is the scale assigned to coordinate $j$.
We compare equivalent parameterizations that give the same current prediction
$f_t$ for the sample.
In that case the loss derivatives $\ell_{f,t}$ and $\ell_{ff,t}$ are unchanged;
only the local output sensitivity changes.
For the rescaled representation,
\[
    \vj_t=\nabla_{\vw}f_{\vw_t}(\tilde{\vx})=\phi\ve_j .
\]
An SGD step is
\[
    \Delta\vw_t^{\mathrm{SGD}}
    =
    -\alpha_t\ell_{f,t}\phi\ve_j,
\]
so its induced first-order prediction change is
\begin{align*}
    h_t^{\mathrm{SGD}}
    =
    \vj_t^\top \Delta\vw_t^{\mathrm{SGD}}
    =
    -\alpha_t\ell_{f,t}\phi^2.
\end{align*}
The effect of SGD on the prediction scales quadratically with the feature
multiplier.

For undamped squared-loss IGND, the same quantity is
\begin{align*}
    h_t^{\mathrm{IGND}}
    =
    -\alpha_t\ell_{f,t}.
\end{align*}
The factor $\phi^2$ cancels: the induced prediction change is independent of
the representation scale.
With Levenberg-Marquardt damping and $\epsilon_c=0$,
\begin{align*}
    h_t^{\mathrm{IGND}}
    =
    -\alpha_t\ell_{f,t}
    \frac{\phi^2}{\phi^2+\epsilon_{\mathrm{LM}}}.
\end{align*}
The invariance is exact without damping and approximate when
$\phi^2\gg\epsilon_{\mathrm{LM}}$.
The sign of $\phi$ is irrelevant 
because the induced prediction change depends on $\phi^2$.
This is a local statement about the first-order prediction change, 
not a global invariance claim for arbitrary nonlinear training dynamics.

\subsection{Comparison with normalized, adaptive, and curvature-based updates}\label{sec:relation-normalized}

We now use the same quantity, $h=\vj^\top\Delta\vw$, 
to compare update rules for one scalar-output sample.
Table~\ref{tab:prediction-space-comparison} shows the parameter update and the
induced first-order prediction change for each method.
The comparison separates methods that control the length of the parameter step
from methods that control the induced change in the scalar prediction.

\begin{table}[t]
\centering
\caption{
Parameter updates and induced first-order prediction changes for one
scalar-output sample. 
Here $\vg=\ell_f\vj$ and $h=\vj^\top\Delta\vw$.
}
\label{tab:prediction-space-comparison}
\begin{tabular}{>{\raggedright\arraybackslash}p{0.20\linewidth} >{\raggedright\arraybackslash}p{0.38\linewidth} >{\raggedright\arraybackslash}p{0.32\linewidth}}
\toprule
Method & Parameter update $\Delta\vw$ & First-order prediction change $h$ \\
\midrule
SGD
& $-\alpha\ell_f\vj$
& $-\alpha\ell_f\norm{\vj}^2$ \\
Normalized gradient
& $-\alpha\vg/(\norm{\vg}+\epsilon)$
& $-\alpha\dfrac{\ell_f\norm{\vj}^2}{|\ell_f|\norm{\vj}+\epsilon}$ \\
SPS 
&
$-\alpha \dfrac{\ell-\ell^\star}{c\norm{\vg}^2}\vg$ 
&
$-\alpha \dfrac{\ell-\ell^\star}{c\ell_f}$ when $\ell_f\ne0$ \\
Undamped IGND
& $-\alpha\dfrac{\ell_f}{\ell_{ff}\norm{\vj}^2}\vj$
& $-\alpha\dfrac{\ell_f}{\ell_{ff}}$ \\
Damped IGND
& $-\alpha\dfrac{\ell_f}{(\ell_{ff}+\epsilon_c)\norm{\vj}^2+\epsilon_{\mathrm{LM}}}\vj$
& $-\alpha\ell_f\dfrac{\norm{\vj}^2}{(\ell_{ff}+\epsilon_c)\norm{\vj}^2+\epsilon_{\mathrm{LM}}}$ \\
Squared-loss IGND / NLMS
& $\alpha\dfrac{r}{\norm{\vj}^2}\vj$, $r=y-f$
& $\alpha r$ \\
\bottomrule
\end{tabular}
\end{table}

\paragraph{SGD and normalized gradient descent}
For SGD, the induced prediction change scales quadratically with the local
sensitivity $\norm{\vj}$.
Normalized gradient descent controls the norm of the parameter step.
For scalar-output models, however, $\norm{\vg}=|\ell_f|\norm{\vj}$; when
$\epsilon=0$, its induced prediction change becomes
$-\alpha\sign(\ell_f)\norm{\vj}$, which still scales linearly with
$\norm{\vj}$~\cite{murray2019revisiting}.
IGND instead normalizes by the rank-one GGN curvature term
$\ell_{ff}\norm{\vj}^2$ and targets the Newton step in the scalar prediction
coordinate in Eq.~\eqref{eq:output-newton-step}.

\paragraph{NLMS}
For squared loss, $\ell_f=f-y$ and $\ell_{ff}=1$. 
If the model is linear and $\epsilon_{\mathrm{LM}}=0$, IGND becomes
\begin{align}
    \vw_{t+1}
    =
    \vw_t
    +
    \alpha_t
    \frac{y_t-f_t}{\norm{\vx_t}^2}
    \vx_t,
\end{align}
which is the usual NLMS update~\cite{widrow1985adaptive,haykin2008adaptive}.
IGND can therefore be viewed as a nonlinear scalar-output GGN extension
of this normalization.

\paragraph{Stochastic Polyak step sizes}
A typical SPS update has the form
\begin{align}\label{eq:sps-generic}
    \vw_{t+1}
    =
    \vw_t
    -
    \alpha_t
    \frac{\ell_t-\ell_t^\star}{c\norm{\vg_t}^2}\vg_t,
\end{align}
possibly with clipping or a maximum step~\cite{loizou2021stochastic,orvieto2022dynamics}.
Its induced first-order prediction change is
$-\alpha_t(\ell_t-\ell_t^\star)/(c\ell_{f,t})$ when $\ell_{f,t}\ne0$.
For squared loss with $\ell_t^\star=0$, 
this equals the IGND prediction
step up to the conventional constants $\alpha_t$ and $c$.
For binary cross-entropy and other non-quadratic losses, 
SPS depends on a loss value ratio, 
whereas IGND depends on the output curvature $\ell_{ff,t}$ 
and the output sensitivity $\norm{\vj_t}^2$.

\paragraph{Adam and diagonal adaptive methods}
AdaGrad, RMSProp, AdaDelta, and Adam use coordinatewise diagonal rescalings
built from accumulated gradient statistics~\cite{duchi2011adaptive,tieleman2012lecture,zeiler2012adadelta,kingma2014adam}.
Their update is not represented by one row of the scalar-rescaling form in
Table~\ref{tab:prediction-space-comparison}: 
the effective preconditioner
depends on the coordinate system and on the optimization history.
This makes Adam-style methods complementary to IGND rather than direct
algebraic equivalents.
The Adam-IGND variant in Section~\ref{sec:experiments} tests this empirical
compatibility.

\paragraph{Gauss-Newton, natural gradient, and mini-batches}
Stochastic Gauss-Newton and natural gradient methods
typically operate with a batch or structured curvature matrix and solve or
approximate a higher-rank linear system~\cite{amari1998natural,martens2020new,gargiani2020promise,ren2019efficient,tran2020stochastic,hong2020stochastic}.
For a mini-batch of $b$ scalar-output samples, 
the GGN matrix has the form
$\mJ^\top\mC\mJ$ and generally has rank at most $b$.
IGND solves the damped rank-one GGN system exactly for each scalar-output sample; 
averaged per-sample directions are practical batched extensions, 
not generally exact mini-batch Gauss-Newton directions.
The Levenberg-Marquardt term in Eq.~\eqref{eq:damped-rank-one-system} is a
scalar-output specialization of classical damped Gauss-Newton
regularization~\cite{levenberg1944method,marquardt1963algorithm,nocedal2006numerical}.

\subsection{Common scalar-output losses and Bellman semi-gradient updates}\label{sec:special-cases}

The previous subsections compared updates 
through the first-order prediction change $h$.
We now return to the parameter update itself.
For a given scalar-output loss, 
IGND is obtained by substituting the output
derivatives $\ell_f$ and $\ell_{ff}$ into Eq.~\eqref{eq:ignd-update}.
For common losses, this gives the following forms.

\paragraph{Squared loss}
For
\begin{align}\label{eq:se}
    \ell^{\mathrm{SE}}(y,f)=\frac{1}{2}(y-f)^2,
\end{align}
we have $\ell_f=f-y$ and $\ell_{ff}=1$.
With $\epsilon_c=0$, Eq.~\eqref{eq:ignd-update} gives
\begin{align}\label{eq:se-update}
    \vw_{t+1}
    =
    \vw_t
    -
    \alpha_t
    \frac{f_t-y_t}{\norm{\vj_t}^2+\epsilon_{\mathrm{LM}}}
    \vj_t.
\end{align}
Equivalently, with residual $r_t=y_t-f_t$,
\begin{align*}
    \vw_{t+1}
    =
    \vw_t
    +
    \alpha_t
    \frac{r_t}{\norm{\vj_t}^2+\epsilon_{\mathrm{LM}}}
    \vj_t.
\end{align*}

\paragraph{Binary cross-entropy with logits}
For binary labels $y\in\{0,1\}$ and logit $f$, let
$p=\sigmoid(f)$, where $\sigmoid(z)=1/(1+\exp(-z))$.
The binary cross-entropy loss is
\begin{align}\label{eq:bce}
    \ell^{\mathrm{BCE}}(y,f)
    =
    -y\log p-(1-y)\log(1-p).
\end{align}
The output derivatives are $\ell_f=p-y$ and $\ell_{ff}=p(1-p)$.
With $p_t=\sigmoid(f_t)$, Eq.~\eqref{eq:ignd-update} gives
\begin{align}\label{eq:bce-update}
    \vw_{t+1}
    =
    \vw_t
    -
    \alpha_t
    \frac{p_t-y_t}
    {\big(p_t(1-p_t)+\epsilon_c\big)\norm{\vj_t}^2+\epsilon_{\mathrm{LM}}}
    \vj_t.
\end{align}
The curvature floor $\epsilon_c$ prevents the output curvature term from
vanishing when $p_t(1-p_t)$ is close to zero, 
while $\epsilon_{\mathrm{LM}}$ provides Levenberg-Marquardt damping.

\paragraph{Bellman error semi-gradient updates}
Let $q_{\vw}(\vs_t,a_t)$ be a scalar action-value estimate, 
and let $y_t$ be a target treated as fixed during differentiation.
For the squared Bellman error
\begin{align*}
    \ell_t
    =
    \frac{1}{2}\big(y_t-q_{\vw_t}(\vs_t,a_t)\big)^2,
\end{align*}
define
\begin{align*}
    \vj_t
    =
    \nabla_{\vw}q_{\vw_t}(\vs_t,a_t),
    \qquad
    \delta_t
    =
    y_t-q_{\vw_t}(\vs_t,a_t).
\end{align*}
Since this is squared loss with prediction $q_{\vw_t}(\vs_t,a_t)$, setting
$\epsilon_c=0$ gives
\begin{align}\label{eq:bellman-ignd-update}
    \vw_{t+1}
    =
    \vw_t
    +
    \alpha_t
    \frac{\delta_t}{\norm{\vj_t}^2+\epsilon_{\mathrm{LM}}}
    \vj_t.
\end{align}
When $y_t$ contains bootstrapped predictions depending on evolving parameters
or policies, Eq.~\eqref{eq:bellman-ignd-update} should be interpreted as a
semi-gradient normalization rather than as the exact gradient step for a fixed
empirical risk objective.

\subsection{Computational cost and batching}\label{sec:cost-batching}

For one scalar-output sample, IGND requires the output sensitivity vector
$\vj_t$, the inner product $\norm{\vj_t}^2$, 
the scalar loss derivatives
$\ell_{f,t}$ and $\ell_{ff,t}$, and one scalar division.
It stores no curvature matrix and solves no linear system.
In reverse-mode automatic differentiation, 
computing $\vj_t$ is comparable to
computing a sample gradient, 
with the additional need to obtain or infer the output sensitivity norm.

The cost model changes for batched implementations.
Averaging IGND directions from individual samples requires 
individual output sensitivity norms.
Depending on the autodiff system, these may require vectorized per-sample
gradients or additional gradient computations, 
and need not have the same cost
as an ordinary mini-batch gradient.
Thus the exact computational claim is the online $b=1$ claim: 
no curvature matrix storage and no linear solve for one scalar-output sample.
Mini-batch variants are implementation choices.

\subsection{Empirical risk stationarity guarantee}\label{sec:convergence}

We next give a stationarity guarantee for damped IGND 
on a fixed empirical risk objective. 
The result shows that the IGND-scaled stochastic direction can be placed inside
the standard smooth nonconvex SGD analysis under explicit alignment and moment
assumptions~\cite{bottou1998online,bottou2018optimization,ghadimi2013stochastic}.
It is scoped to a fixed empirical risk objective with i.i.d. sampling;
it does not prove convergence of bootstrapped reinforcement learning dynamics.

Let $\zeta=(\vx,y)$ 
denote a sample and define the expected objective
\begin{align}\label{eq:expected-objective}
    \bar{\Ls}(\vw)
    :=
    \E_{\zeta}\big[\ell(y,f_{\vw}(\vx))\big].
\end{align}
When $\zeta$ is sampled uniformly from a finite dataset $\mathcal{D}$, 
this is the empirical risk in Eq.~\eqref{eq:problem}.
Let $\zeta_t$ be sampled i.i.d. and let
$\cF_t=\sigma(\vw_0,\zeta_0,\ldots,\zeta_{t-1})$ denote the history before
iteration $t$.
We write $\E_t[\cdot]=\E[\cdot\mid\cF_t]$.
Define
\begin{align}\label{eq:convergence-quantities}
    \vg_t
    &:= \nabla_{\vw}\ell(y_t,f_{\vw_t}(\vx_t)),
    &
    \bar{\vg}_t
    &:= \nabla\bar{\Ls}(\vw_t)
      = \E_t[\vg_t],
    \nonumber\\
    \widetilde{\vg}_t
    &:= \xi_t\vg_t,
    &
    \vw_{t+1}
    &:= \vw_t-\alpha_t\widetilde{\vg}_t.
\end{align}
The scalar $\xi_t$ is neither deterministic 
nor independent of the stochastic gradient: 
both $\xi_t$ and $\vg_t$ are computed from the same sampled pair
through $\ell_{ff,t}$, $\ell_{f,t}$, and $\vj_t$.
The analysis therefore works directly with the scaled stochastic gradient
$\widetilde{\vg}_t=\xi_t\vg_t$, 
rather than factorizing $\E_t[\xi_t\vg_t]$.
Assumption~\ref{ass:alignment-moment} is stated directly for 
$\widetilde{\vg}_t$.
The proof in~\ref{apx:conv-proof} gives one conservative sufficient condition,
based on bounded sensitivity, bounded output curvature, and strong growth,
under which this assumption holds.

\begin{assumption}[Empirical risk regularity]\label{ass:empirical-risk-regularity}
The samples $\zeta_t$ are i.i.d. and independent of $\cF_t$.
The objective $\bar{\Ls}$ is bounded below by $\bar{\Ls}_{\inf}>-\infty$ and
has an $L$-Lipschitz gradient:
\begin{align*}
    \norm{\nabla\bar{\Ls}(\vw)-\nabla\bar{\Ls}(\vu)}
    \le
    L\norm{\vw-\vu}
    \qquad \text{for all } \vw,\vu.
\end{align*}
The damping parameters are fixed with $\epsilon_c\ge0$ and
$\epsilon_{\mathrm{LM}}>0$.
\end{assumption}

\begin{assumption}[Alignment and second moment]\label{ass:alignment-moment}
There exist constants $\mu>0$, $M_0<\infty$, and $M_2<\infty$ such that, 
for all $t$,
\begin{align}
    \bar{\vg}_t^\top \E_t[\widetilde{\vg}_t]
    &\ge
    \mu\norm{\bar{\vg}_t}^2,
    \label{eq:main-alignment-assumption}
    \\
    \E_t\big[\norm{\widetilde{\vg}_t}^2\big]
    &\le
    M_0+M_2\norm{\bar{\vg}_t}^2.
    \label{eq:main-moment-assumption}
\end{align}
\end{assumption}

Assumption~\ref{ass:alignment-moment} is the IGND-specific part of the
analysis.
It says that the expected scaled stochastic gradient keeps a positive component
along the full gradient and has an affine conditional second-moment bound.
The assumption is stated for $\widetilde{\vg}_t$ itself because $\xi_t$ and
$\vg_t$ are generally dependent.
The sufficient condition in~\ref{apx:conv-proof} is conservative and is not a
tuning prescription.

\begin{assumption}[Robbins-Monro step sizes]\label{ass:robbins-monro}
The learning rate sequence satisfies
\begin{align}
    \sum_{t=0}^{\infty}\alpha_t=\infty,
    \qquad
    \sum_{t=0}^{\infty}\alpha_t^2<\infty.
\end{align}
\end{assumption}

\begin{theorem}[Stationarity under aligned IGND scaling]\label{thm:ignd-conv}
Suppose Assumptions~\ref{ass:empirical-risk-regularity},
\ref{ass:alignment-moment}, and~\ref{ass:robbins-monro} hold.
Let $\{\vw_t\}$ be generated by damped IGND.
Then
\begin{align}\label{eq:weighted-sum-stationarity}
    \sum_{t=0}^{\infty}
    \alpha_t
    \E\big[\norm{\nabla\bar{\Ls}(\vw_t)}^2\big]
    <
    \infty.
\end{align}
Consequently,
\begin{align}\label{eq:weighted-average-stationarity}
    \lim_{T\to\infty}
    \E\left[
    \frac{\sum_{t=0}^{T}\alpha_t\norm{\nabla\bar{\Ls}(\vw_t)}^2}
    {\sum_{t=0}^{T}\alpha_t}
    \right]
    =0,
\end{align}
and
\begin{align}\label{eq:liminf-stationarity}
    \liminf_{t\to\infty}
    \E\big[\norm{\nabla\bar{\Ls}(\vw_t)}^2\big]
    =0.
\end{align}
\end{theorem}

The proof follows the usual smoothness-based stochastic approximation template
for nonconvex SGD, 
applied to the scaled stochastic gradient $\widetilde{\vg}_t=\xi_t\vg_t$.
The affine moment bound in Eq.~\eqref{eq:main-moment-assumption} is sufficient
because $\sum_t\alpha_t^2<\infty$.
The full proof is in~\ref{apx:conv-proof}.
Thus Theorem~\ref{thm:ignd-conv} should be read as a conditional stationarity
guarantee under 
Assumptions~\ref{ass:empirical-risk-regularity}--\ref{ass:robbins-monro},
not as an unconditional convergence result for arbitrary models, 
sampling distributions, or bootstrapped learning dynamics.

\section{Experiments}\label{sec:experiments}

The experiments evaluate the scalar-output normalization in three settings.
The FrozenLake scale test checks the first-order prediction change behavior
derived in Section~\ref{sec:scale}. 
The supervised experiments test IGND on
standard regression and binary classification tasks with scalar outputs. 
The LQR case study applies the squared Bellman error version of the update to
learning an action-value function inside policy iteration.

\subsection{Experimental protocol}\label{sec:exp-protocol}

All curves report the mean over $20$ random seeds, 
with shaded regions showing $\pm 1$ standard deviation. 
Optimizer learning rates are selected using validation performance, 
and final metrics are reported on held-out test or
evaluation data. 
Test and evaluation data are not used for learning rate selection. 
The learning rate search uses logarithmic candidates spanning
$10^{-9}$ to $1$. 
For Adam and Adam-IGND, only the learning rate is selected from this grid. 
The Adam moment parameters and numerical constant are
fixed to the same values for both methods.

The damping parameters are fixed by loss family in all experiments:
$\epsilon_c=0$ and $\epsilon_{\mathrm{LM}}=10^{-5}$ for squared error losses,
and $\epsilon_c=10^{-2}$ and $\epsilon_{\mathrm{LM}}=10^{-5}$ for binary
cross-entropy. 
Since the exact rank-one GGN interpretation applies to one
sample at a time, all experiments use batch size $1$. 
Architecture, preprocessing, metric definitions, optimizer constants, 
and LQR details are given in~\ref{apx:exp_details}.

\subsection{Controlled scale robustness}\label{sec:exp-scale-robustness}

We use \texttt{FrozenLake-v1}~\cite{openai_gym} 
to build a scale robustness test in which the task stays fixed 
and only the feature representation changes.
The Q-value function is tabular and one-hot, 
so multiplying feature coordinates
changes the local output sensitivity without changing the underlying task.

\begin{figure*}[t]
    \centering
    \includegraphics[width=0.98\textwidth]{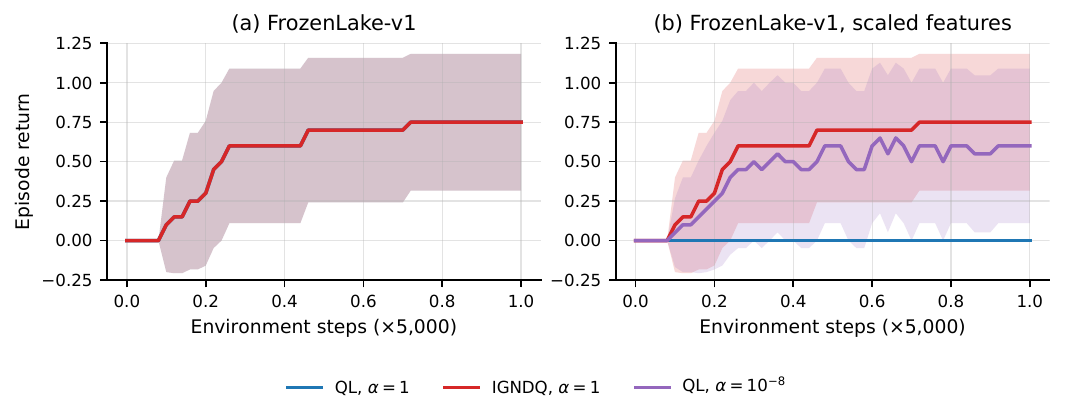}
    \caption{Controlled feature scaling on \texttt{FrozenLake-v1}. Curves show
    mean episode return over $20$ seeds, with shaded $\pm 1$ standard deviation.}
    \label{fig:fl_results}
\end{figure*}

This is the one-hot scaling case analyzed in Section~\ref{sec:scale}. 
For a given update, 
let $\phi$ denote the scale assigned to the active one-hot coordinate.
Q-learning (QL) with a squared Bellman
error changes the current prediction by an amount proportional to $\phi^2$.
Undamped IGNDQ gives
$h_t=\vj_t^\top\Delta\vw_t=-\alpha_t\ell_{f,t}$, independent of $\phi$.
Thus the normalization removes the $\phi^2$ sensitivity factor 
from the first-order prediction change, 
rather than merely shrinking the parameter step.

Figure~\ref{fig:fl_results} supports the scale robustness claim. 
With the original one-hot features, 
QL and IGNDQ with $\alpha=1$ behave similarly. 
In the scaled feature condition, 
each one-hot coordinate is multiplied by a fixed
random nonzero integer in $[-10^4,10^4]$, with the same scaling used for all
methods and seeds. 
After this rescaling, 
QL with $\alpha=1$ does not learn within the plotted horizon 
because the update changes the current prediction
in proportion to the squared feature multiplier. 
IGNDQ with the same learning rate remains stable 
because its denominator contains $\norm{\vj_t}^2$,
which equals $\phi^2$ in this one-hot case.
This cancels the squared feature multiplier in $h_t$.
The additional QL curve with $\alpha=10^{-8}$ shows that retuning can partly
compensate for the changed sensitivity.

\subsection{Supervised learning}\label{sec:exp_sl}

\begin{figure*}[t]
    \centering
    \includegraphics[width=0.98\textwidth]{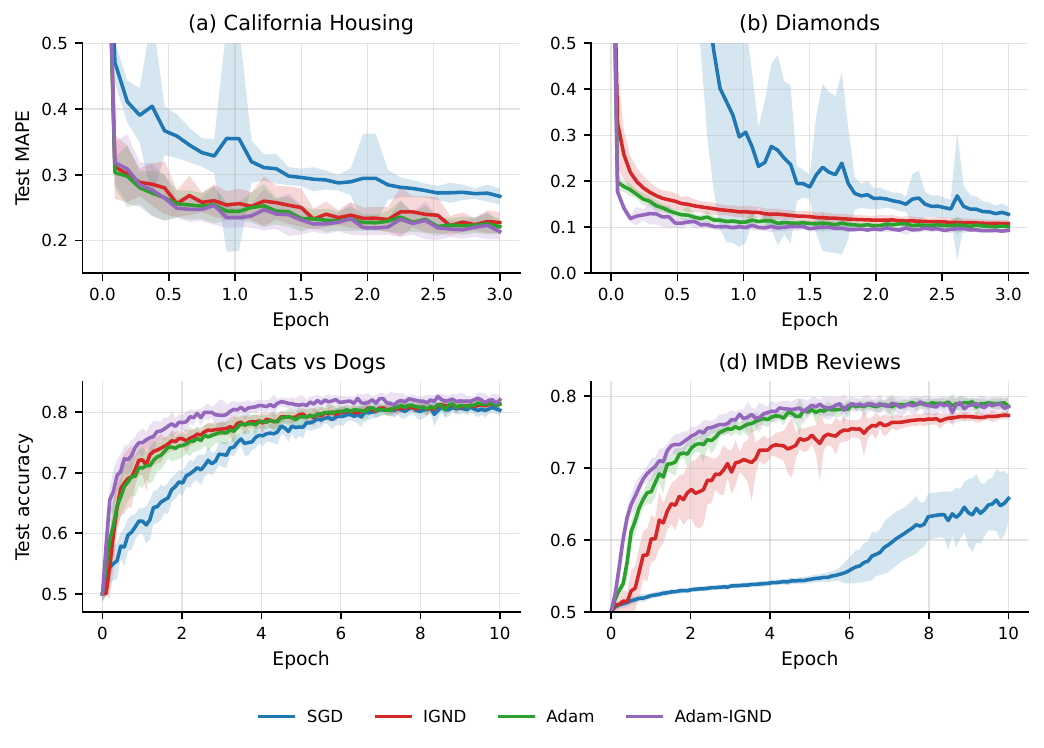}
    \caption{
    Learning curves on the supervised scalar-output tasks. Regression
    panels report MAPE; binary classification panels report test accuracy.
    }
    \label{fig:sl_new}
\end{figure*}

\begin{table*}[t]
\centering
\small
\setlength{\tabcolsep}{4.25pt}
\renewcommand{\arraystretch}{1.05}
\caption{
Final performance on the test set.
Values are mean $\pm 1$ standard deviation over $20$ seeds; 
regression tasks report MAPE and classification tasks report accuracy. 
Bold denotes the best displayed mean in each column.
}
\label{tab:perf_sl}
\begin{tabular}{@{}lcccc@{}}
\toprule
\textbf{Optimizer}
& \textbf{\shortstack{California\\Housing}}
& \textbf{\shortstack{Diamonds}}
& \textbf{\shortstack{Cats vs\\Dogs}}
& \textbf{\shortstack{IMDB\\Reviews}} \\
\midrule
SGD
& 0.267 $\pm$ 0.012
& 0.128 $\pm$ 0.017
& 0.803 $\pm$ 0.010
& 0.658 $\pm$ 0.033 \\
IGND
& 0.227 $\pm$ 0.016
& 0.108 $\pm$ 0.007
& 0.813 $\pm$ 0.007
& 0.773 $\pm$ 0.006 \\
Adam
& 0.221 $\pm$ 0.012
& 0.101 $\pm$ 0.004
& 0.813 $\pm$ 0.007
& \textbf{0.786 $\pm$ 0.015} \\
Adam-IGND
& \textbf{0.213 $\pm$ 0.012}
& \textbf{0.093 $\pm$ 0.005}
& \textbf{0.821 $\pm$ 0.008}
& \textbf{0.786 $\pm$ 0.012} \\
\bottomrule
\end{tabular}
\end{table*}

We next evaluate IGND on standard supervised tasks with scalar outputs. 
The regression tasks are California Housing~\cite{ds_california_housing} and
Diamonds~\cite{ds_diamonds}, trained with squared loss. 
The binary classification tasks are Cats vs Dogs~\cite{ds_cats_vs_dogs} and IMDB
Reviews~\cite{ds_imdb_reviews}, 
trained with binary cross-entropy with logits.

We compare SGD, IGND, Adam, and an exploratory variant called Adam-IGND. 
For Adam-IGND, we first compute the direction scaled by IGND 
and then feed that direction into the same update used by the Adam baseline.
Adam-IGND is included to test whether the scalar GGN normalizer can be paired
with Adam-style adaptive updates.
It is not part of the main derivation or the convergence theorem.

Figure~\ref{fig:sl_new} and Table~\ref{tab:perf_sl} show 
that standalone IGND improves over SGD in the final mean metric on all four tasks. 
Compared with Adam, 
standalone IGND is competitive rather than uniformly better: 
Adam has better mean final performance on California Housing, Diamonds, and IMDB
Reviews, and the same displayed mean on Cats vs Dogs. 
Adam-IGND matches or improves Adam's displayed mean on all four tasks.

Taken together, these results show that the scalar GGN normalization performs
well as a standalone update on these scalar-output tasks and can also serve as
a useful component of adaptive first-order methods.

\subsection{Linear-quadratic control}\label{sec:exp_lqr}

\begin{figure*}[t]
    \centering
    \includegraphics[width=0.98\textwidth]{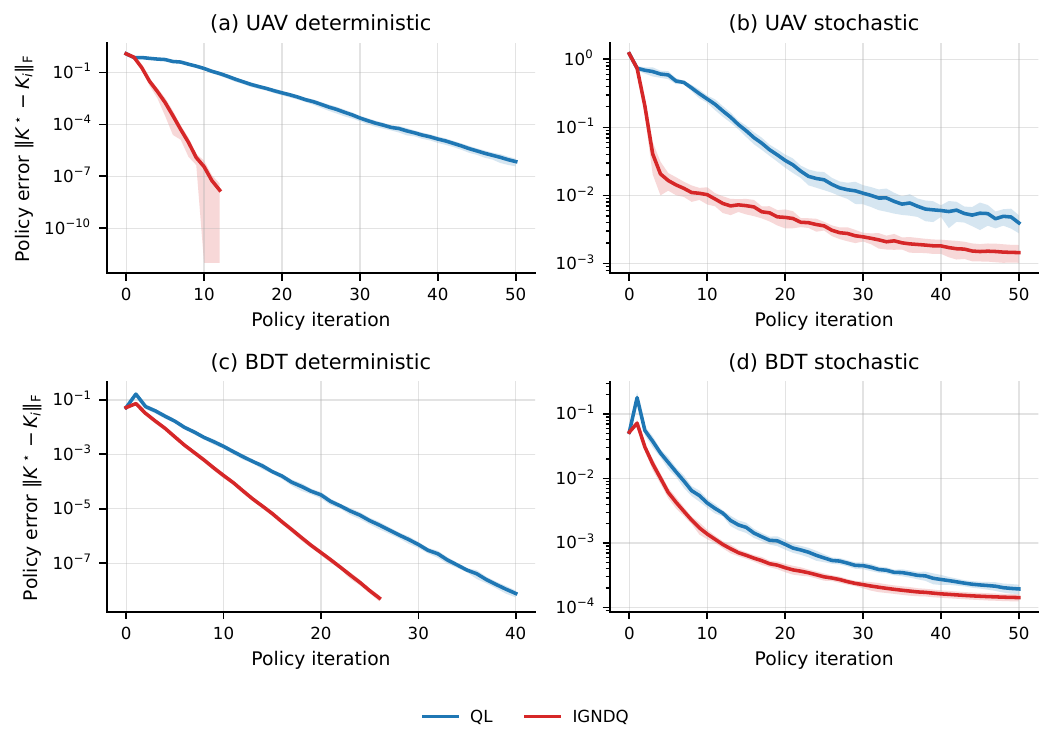}
    \caption{
    LQR case study. Curves show mean Frobenius policy error
    $\norm{\mK^\star-\mK_i}_{F}$ on deterministic and stochastic
    variants of the UAV and BDT systems.
    }
    \label{fig:lqr}
\end{figure*}

Finally, we test the Bellman error version of IGND in a controlled
linear-quadratic setting. 
Following~\cite{bradtke1994adaptive,bradtke1996linear},
policy evaluation learns a quadratic action-value function 
and policy improvement forms the greedy linear feedback policy. 
We compare the standard Q-learning semi-gradient update, denoted QL, 
with the corresponding IGNDQ update. 
In both methods, the Bellman target is treated as fixed during each
policy evaluation update.

The learned feedback matrix $\mK_i$ is compared with the optimal LQR policy
$\mK^\star$ on a linearized UAV model~\cite{hung1982multivariable} 
and a binary distillation tower model~\cite{davison1990benchmark}. 
Details are given in~\ref{apx:exp_details}.

Figure~\ref{fig:lqr} shows that IGNDQ reduces policy error faster than QL 
in these LQR experiments. 
The advantage is clearest in the deterministic panels,
where IGNDQ reaches very small policy error in fewer recorded policy improvement iterations.
The shorter deterministic IGNDQ curves reflect the number of recorded policy improvement iterations; 
no padding or extrapolation is applied. 
In the stochastic panels, IGNDQ also maintains lower error over most of training.

These experiments test the same semi-gradient normalization in a controlled
setting where the learned object is an action-value function.
Theorem~\ref{thm:ignd-conv} gives a stationarity guarantee for fixed i.i.d.
empirical risk objectives; 
the LQR study instead tests the update inside bootstrapped policy iteration.

Overall, the experiments support the empirical claim: 
the rank-one scalar normalization is robust to changes in local output sensitivity, 
performs well on the tested supervised tasks with scalar outputs, 
and improves the tested semi-gradient action-value updates. 
Extensions to vector outputs, exact
mini-batch Gauss-Newton systems, and broader reinforcement learning settings
require separate treatment.

\section{Discussion and Limitations}\label{sec:discussion}

\paragraph{Where the method is strongest}
IGND is most relevant when the sensitivity of a scalar model output varies
substantially across samples, features, states, or training iterations. 
In this regime, a fixed SGD learning rate can be limited by the largest sensitivities,
because those samples produce large changes in the current prediction. 
IGND rescales the sample gradient using the current sensitivity 
$\norm{\vj_t}^2$ and output curvature $\ell_{ff,t}$, 
so the update is controlled in the local scalar-output coordinate rather 
than only by the size of the parameter step. 
This explains the scale robustness test and is consistent with the LQR experiments.

\paragraph{Scope of the derivation}
The exact IGND derivation applies to one scalar-output sample. 
Losses with vector outputs, multiclass softmax losses, 
and true mini-batch GGN systems generally produce higher-rank curvature matrices. 
Extending the same idea to those settings therefore requires a separate derivation. 
IGND should be read as complementary to Adam, KFAC, Hessian-free optimization, 
and mini-batch Gauss-Newton methods, 
not as a replacement for them.

\paragraph{Batched implementation}
This paper uses batch size $1$ to match the exact rank-one derivation. 
A batched implementation could average IGND directions computed from individual samples, 
but that average is not generally the exact mini-batch Gauss-Newton direction. 
It also requires per-sample sensitivity norms, 
which may be more expensive than an ordinary mini-batch gradient depending on the autodiff implementation.

\paragraph{Value learning scope}
The Bellman error update in Eq.~\eqref{eq:bellman-ignd-update} is exact as a
gradient step when the target is fixed. 
With bootstrapped targets, changing replay distributions, 
target networks, or policy-dependent data, it should be interpreted as a semi-gradient normalization. 
The LQR experiments are therefore a controlled value learning case study of the update. 
They are not a consequence of Theorem~\ref{thm:ignd-conv}, 
which applies to fixed i.i.d. empirical risk objectives.

\paragraph{Hybrid adaptive variants}
The Adam-IGND results suggest that the scalar GGN normalizer can be combined
with Adam-style adaptive updates. 
One option is to pass the IGND-scaled sample gradient to the same Adam
update equations used by the baseline; 
the exact update is given in~\ref{apx:exp_details}.
\section{Conclusions and Future Work}\label{sec:outro}

This paper developed IGND from a structural property of scalar-output losses
evaluated one sample at a time. 
For one such sample, the GGN matrix has rank at most one; 
when the sample gradient is nonzero, 
both the gradient and the only possible nonzero GGN curvature direction 
lie in the output-sensitivity span.
This makes the Moore-Penrose and damped Gauss-Newton directions
available in closed form as scalar normalizations of the sample gradient. 
The resulting update requires no curvature matrix storage, factorization, 
or iterative linear solve.

The first-order prediction change $h_t=\vj_t^\top\Delta\vw_t$ explains what
the normalization controls. 
IGND is not just shortening the parameter step. 
In the undamped case, it produces the scalar Newton change
$h_t=-\alpha_t\ell_{f,t}/\ell_{ff,t}$ for the local quadratic loss model.
The Moore-Penrose construction then selects the minimum-norm parameter
displacement that realizes this change. 
This distinguishes IGND from methods that normalize only the parameter step, 
and explains its relation to NLMS for
squared loss and to SPS under quadratic losses.

The experiments support the practical value of this normalization on scalar-output supervised learning, 
a controlled scale robustness test, 
and an LQR case study. 
Natural next steps are principled extensions
to mini-batches and vector outputs, tighter theory for bootstrapped
semi-gradient value learning, 
and a systematic study of Adam-style combinations
with scalar-output GGN normalization.

\bibliographystyle{plainnat}
\bibliography{references}

\appendix

\section{Additional derivation details}\label{apx:derivation}

\subsection{Moore-Penrose pseudoinverse details}
\label{apx:mp-rank-one}

For completeness, 
we recall the rank-one Moore-Penrose pseudoinverse formula
used in the IGND derivation. 
If $c \neq 0$ and $\norm{\vj}>0$, 
then
\begin{align*}
    (c\vj\vj^\top)^\dagger
    =
    \frac{1}{c}\frac{\vj\vj^\top}{\norm{\vj}^2 \norm{\vj}^2}
    =
    \frac{1}{c}\frac{\vj\vj^\top}{\norm{\vj}^4}.
\end{align*}
Substituting $c=\ell_{ff,t}$ gives Eq.~\eqref{eq:rank-one-pinv}. 
This identity is the only generalized inverse fact needed for the main
derivation. 
The next subsection gives an equivalent quadratic programming
view of the same undamped one-sample GGN direction.

\subsection{Quadratic programming view of the IGND update}
\label{apx:qp-rank-one}

The Moore-Penrose direction in Eq.~\eqref{eq:mp-ggn-direction} 
can also be obtained by explicitly removing 
the null-space component of the one-sample GGN quadratic model. 
This is only a geometric characterization of the undamped rank-one direction; 
Algorithm~\ref{alg:ignd} uses the damped closed form in Eq.~\eqref{eq:xi-reg}.

\begin{proposition}[Geometric form of the IGND update]
\label{prop:qp-rank-one-collapse}
Assume $\ell_{ff,t}>0$ and $\norm{\vj_t}>0$. 
Let $\mZ_t\in\R^{d\times(d-1)}$ have orthonormal columns spanning the null space
of $\vj_t^\top$, 
such that
\[
    \vj_t^\top\mZ_t=\vzero^\top,
    \qquad
    \mZ_t^\top\mZ_t=\mI .
\]
Consider the singular rank-one GGN quadratic problem
\begin{align}
    \min_{\Delta\vw\in\R^d}\;
    \frac{1}{2}\ell_{ff,t}(\vj_t^\top\Delta\vw)^2
    +\ell_{f,t}\vj_t^\top\Delta\vw .
    \label{eq:rank-one-ggn-qp}
\end{align}
The auxiliary null-space-penalized problem
\begin{align}
    \widetilde{\vd}_t
    :=
    \argmin_{\Delta\vw\in\R^d}\;
    \frac{1}{2}\ell_{ff,t}(\vj_t^\top\Delta\vw)^2
    +\ell_{f,t}\vj_t^\top\Delta\vw
    +\frac{1}{2}\norm{\mZ_t^\top\Delta\vw}^2
    \label{eq:rank-one-ggn-qp-aux}
\end{align}
has the unique solution
\begin{align}
    \widetilde{\vd}_t
    =
    -
    \frac{\ell_{f,t}}
    {\ell_{ff,t}\norm{\vj_t}^2}
    \vj_t .
    \label{eq:rank-one-ggn-qp-solution}
\end{align}
Moreover, this direction is the unique minimum-norm minimizer of
Eq.~\eqref{eq:rank-one-ggn-qp}. 
Hence
$\widetilde{\vd}_t=\vd_t^{\mathrm{GGN}}$ as defined in
Eq.~\eqref{eq:mp-ggn-direction}.
\end{proposition}

\begin{proof}
The first-order optimality condition for
Eq.~\eqref{eq:rank-one-ggn-qp-aux} is
\begin{align}
    \ell_{ff,t}\vj_t(\vj_t^\top\Delta\vw)
    +\ell_{f,t}\vj_t
    +\mZ_t\mZ_t^\top\Delta\vw
    =
    \vzero .
    \label{eq:rank-one-ggn-qp-aux-foc}
\end{align}
Let
\[
    \vu_t:=\frac{\vj_t}{\norm{\vj_t}},
\]
and decompose
\[
    \Delta\vw
    =
    \vu_t\,\Delta w^{\parallel}
    +
    \mZ_t\Delta\vw^{\perp},
\]
where $\Delta w^{\parallel}\in\R$ and
$\Delta\vw^{\perp}\in\R^{d-1}$. 
Since $\mZ_t^\top\vu_t=\vzero$ and $\mZ_t^\top\mZ_t=\mI$, 
substituting this
decomposition into Eq.~\eqref{eq:rank-one-ggn-qp-aux-foc} gives
\[
    \ell_{ff,t}\vj_t
    \big(\vj_t^\top\vu_t\,\Delta w^{\parallel}\big)
    +\ell_{f,t}\vj_t
    +\mZ_t\Delta\vw^{\perp}
    =
    \vzero .
\]
Premultiplying by $\mZ_t^\top$ yields $\Delta\vw^{\perp}=\vzero$. 
Thus the auxiliary problem removes the null-space
component and its solution lies in $\mathrm{span}\{\vj_t\}$. 
Since $\vj_t^\top\vu_t=\norm{\vj_t}$, 
the remaining optimality condition is
\[
    \ell_{ff,t}\norm{\vj_t}\Delta w^{\parallel}
    +\ell_{f,t}
    =
    0,
\]
so
\[
    \Delta w^{\parallel}
    =
    -
    \frac{\ell_{f,t}}
    {\ell_{ff,t}\norm{\vj_t}} .
\]
Therefore
\[
    \Delta\vw
    =
    -
    \frac{\ell_{f,t}}
    {\ell_{ff,t}\norm{\vj_t}^2}
    \vj_t ,
\]
which proves Eq.~\eqref{eq:rank-one-ggn-qp-solution}.

It remains to relate this direction to the singular problem in
Eq.~\eqref{eq:rank-one-ggn-qp}. 
Its first-order condition is
\[
    \ell_{ff,t}\vj_t(\vj_t^\top\Delta\vw)
    +\ell_{f,t}\vj_t
    =
    \vzero ,
\]
and, since $\vj_t\neq\vzero$, every minimizer satisfies
\[
    \vj_t^\top\Delta\vw
    =
    -
    \frac{\ell_{f,t}}{\ell_{ff,t}} .
\]
Thus every minimizer can be written as
\[
    \Delta\vw
    =
    \widetilde{\vd}_t+\vz,
    \qquad
    \vj_t^\top\vz=0 .
\]
Because $\widetilde{\vd}_t\in\mathrm{span}\{\vj_t\}$, 
this decomposition is orthogonal, and hence
\[
    \norm{\Delta\vw}^2
    =
    \norm{\widetilde{\vd}_t}^2+\norm{\vz}^2 .
\]
The norm is minimized uniquely when $\vz=\vzero$. 
Therefore $\widetilde{\vd}_t$ is the unique minimum-norm minimizer 
of the singular rank-one quadratic problem, 
which is exactly the Moore-Penrose direction in
Eq.~\eqref{eq:mp-ggn-direction}.
\end{proof}

\section{Full convergence proof}\label{apx:conv-proof}

This appendix proves Theorem~\ref{thm:ignd-conv} and gives one sufficient set of
conditions under which Assumption~\ref{ass:alignment-moment} holds. 
All expectations below are conditional on $\cF_t$ unless stated otherwise.

\begin{lemma}[One-step descent]\label{lem:one-step-descent-appendix}
Under Assumption~\ref{ass:empirical-risk-regularity}, the damped IGND update
$\vw_{t+1}=\vw_t-\alpha_t\widetilde{\vg}_t$ satisfies
\begin{align}
    \E_t[\bar{\Ls}(\vw_{t+1})]-\bar{\Ls}(\vw_t)
    \le
    -\alpha_t\bar{\vg}_t^\top\E_t[\widetilde{\vg}_t]
    +
    \frac{L}{2}\alpha_t^2\E_t\big[\norm{\widetilde{\vg}_t}^2\big].
    \label{eq:appendix-one-step-descent}
\end{align}
If Assumption~\ref{ass:alignment-moment} also holds, then
\begin{align}
    \E_t[\bar{\Ls}(\vw_{t+1})]-\bar{\Ls}(\vw_t)
    \le
    -\left(\mu-\frac{L}{2}\alpha_tM_2\right)
    \alpha_t\norm{\bar{\vg}_t}^2
    +
    \frac{L}{2}M_0\alpha_t^2.
    \label{eq:appendix-expected-decrease}
\end{align}
\end{lemma}

\begin{proof}
By $L$-smoothness,
\begin{align*}
    \bar{\Ls}(\vw_{t+1})
    \le
    \bar{\Ls}(\vw_t)
    +
    \nabla\bar{\Ls}(\vw_t)^\top(\vw_{t+1}-\vw_t)
    +
    \frac{L}{2}\norm{\vw_{t+1}-\vw_t}^2.
\end{align*}
Using $\vw_{t+1}-\vw_t=-\alpha_t\widetilde{\vg}_t$ and conditioning on $\cF_t$ gives
\eqref{eq:appendix-one-step-descent}. Substituting
\eqref{eq:main-alignment-assumption} and~\eqref{eq:main-moment-assumption}
gives
\begin{align*}
    \E_t[\bar{\Ls}(\vw_{t+1})]-\bar{\Ls}(\vw_t)
    &\le
    -\alpha_t\mu\norm{\bar{\vg}_t}^2
    +
    \frac{L}{2}\alpha_t^2\left(M_0+M_2\norm{\bar{\vg}_t}^2\right)\\
    &=
    -\left(\mu-\frac{L}{2}\alpha_tM_2\right)
    \alpha_t\norm{\bar{\vg}_t}^2
    +
    \frac{L}{2}M_0\alpha_t^2,
\end{align*}
which proves~\eqref{eq:appendix-expected-decrease}.
\end{proof}

\begin{proof}[Proof of Theorem~\ref{thm:ignd-conv}]
By Assumption~\ref{ass:robbins-monro}, $\alpha_t\to0$. 
Hence there exists $T_0$ such that, for all $t\ge T_0$,
\begin{align*}
    \frac{L}{2}\alpha_t M_2 \le \frac{\mu}{2}.
\end{align*}
Therefore,
\begin{align*}
    \mu-\frac{L}{2}\alpha_tM_2
    \ge
    \frac{\mu}{2}.
\end{align*}
Taking total expectations in~\eqref{eq:appendix-expected-decrease} and summing
from $t=T_0$ to $T$ yields
\begin{align*}
    \frac{\mu}{2}
    \sum_{t=T_0}^{T}
    \alpha_t\E\big[\norm{\nabla\bar{\Ls}(\vw_t)}^2\big]
    &\le
    \E[\bar{\Ls}(\vw_{T_0})]-\E[\bar{\Ls}(\vw_{T+1})]
    +
    \frac{L}{2}M_0\sum_{t=T_0}^{T}\alpha_t^2
    \\
    &\le
    \E[\bar{\Ls}(\vw_{T_0})]-\bar{\Ls}_{\inf}
    +
    \frac{L}{2}M_0\sum_{t=T_0}^{\infty}\alpha_t^2.
\end{align*}
Letting $T\to\infty$ gives~\eqref{eq:weighted-sum-stationarity}; the finitely
many terms before $T_0$ do not affect finiteness. Dividing the finite weighted
sum by $A_T=\sum_{t=0}^{T}\alpha_t$ and using $A_T\to\infty$ gives
\eqref{eq:weighted-average-stationarity}. 
Finally, if
$\liminf_t\E[\norm{\nabla\bar{\Ls}(\vw_t)}^2]>0$, 
then the divergent series $\sum_t\alpha_t$ would force
$\sum_t\alpha_t\E[\norm{\nabla\bar{\Ls}(\vw_t)}^2]=\infty$, 
contradicting~\eqref{eq:weighted-sum-stationarity}. 
Therefore \eqref{eq:liminf-stationarity} holds.
\end{proof}

\subsection*{A sufficient condition for the alignment and moment bounds}

The main theorem assumes the alignment and moment conditions directly. 
The next result records one conservative sufficient condition. 
It is useful for showing that 
the damped scalar normalization can be placed inside a standard SGD stationarity proof, 
but the constants should not be read as practical tuning rules.

\begin{assumption}[Bounded output sensitivity and curvature]\label{ass:bounded-sensitivity-curvature}
Along the iterates, there are constants $Q<\infty$ and
$\Lambda_{ff}<\infty$ such that, conditionally on $\cF_t$,
\begin{align*}
    \norm{\vj_t}^2\le Q,
    \qquad
    0\le \ell_{ff,t}\le \Lambda_{ff}
\end{align*}
for $\zeta=(\vx,y)$ almost surely, for all $t$.
\end{assumption}

\begin{assumption}[Strong growth]\label{ass:strong-growth-appendix}
There exists $\rho\ge1$ such that, for all $t$,
\begin{align*}
    \E_t\big[\norm{\vg_t}^2\big]
    \le
    \rho\norm{\bar{\vg}_t}^2.
\end{align*}
\end{assumption}
Strong-growth conditions are used in analyses of stochastic gradient methods
under interpolation and overparameterization~\cite{schmidt2013fast,vaswani2019fast}.

\begin{lemma}[Bounded damped scalar]\label{lem:bounded-xi-appendix}
Under Assumption~\ref{ass:bounded-sensitivity-curvature}, the damped IGND
scalar satisfies
\begin{align*}
    \xi_{\min}\le \xi_t\le \xi_{\max},
\end{align*}
where
\begin{align*}
    \xi_{\min}
    :=
    \frac{1}{(\Lambda_{ff}+\epsilon_c)Q+\epsilon_{\mathrm{LM}}},
    \qquad
    \xi_{\max}
    :=
    \frac{1}{\epsilon_{\mathrm{LM}}}.
\end{align*}
\end{lemma}

\begin{proof}
The denominator of $\xi_t$ satisfies
\begin{align*}
    \epsilon_{\mathrm{LM}}
    \le
    (\ell_{ff,t}+\epsilon_c)\norm{\vj_t}^2+\epsilon_{\mathrm{LM}}
    \le
    (\Lambda_{ff}+\epsilon_c)Q+\epsilon_{\mathrm{LM}}.
\end{align*}
Taking reciprocals gives the result.
\end{proof}

\begin{proposition}[Sufficient alignment and moment bounds]\label{prop:sufficient-alignment-moment}
Suppose Assumptions~\ref{ass:bounded-sensitivity-curvature} and
\ref{ass:strong-growth-appendix} hold. Let
\begin{align*}
    \bar{\sigma}_{\xi}
    &:=
    \frac{1}{2}(\xi_{\max}-\xi_{\min}),
    &
    \mu
    &:=
    \xi_{\min}-\bar{\sigma}_{\xi}\sqrt{\rho-1},
    \\
    M_2
    &:=
    \xi_{\max}^2\rho,
    &
    M_0
    &:=
    0.
\end{align*}
If $\mu>0$, then Assumption~\ref{ass:alignment-moment} holds with these
constants.
\end{proposition}

\begin{proof}
All variances below are conditional on $\cF_t$.
For vectors, we write
\[
    \Var_t[\vg_t]
    :=
    \E_t\big[\norm{\vg_t-\bar{\vg}_t}^2\big].
\]
First, Lemma~\ref{lem:bounded-xi-appendix} gives
$\xi_t\in[\xi_{\min},\xi_{\max}]$ almost surely.
By Popoviciu's inequality for bounded random variables
\cite{popoviciu1935equations,sharma2010some},
\begin{align*}
    \Var_t[\xi_t]
    \le
    \frac{1}{4}(\xi_{\max}-\xi_{\min})^2
    =
    \bar{\sigma}_{\xi}^2.
\end{align*}
Strong growth implies
\begin{align*}
    \Var_t[\vg_t]
    :=
    \E_t\big[\norm{\vg_t-\bar{\vg}_t}^2\big]
    \le
    (\rho-1)\norm{\bar{\vg}_t}^2.
\end{align*}
Using the covariance decomposition,
\begin{align*}
    \E_t[\xi_t\vg_t]
    =
    \E_t[\xi_t]\bar{\vg}_t
    +
    \Cov_t(\xi_t,\vg_t),
\end{align*}
where
\begin{align*}
    \Cov_t(\xi_t,\vg_t)
    :=
    \E_t\big[(\xi_t-\E_t[\xi_t])(\vg_t-\bar{\vg}_t)\big].
\end{align*}
Cauchy-Schwarz gives
\begin{align*}
    \norm{\Cov_t(\xi_t,\vg_t)}
    \le
    \sqrt{\Var_t[\xi_t]}\sqrt{\Var_t[\vg_t]}
    \le
    \bar{\sigma}_{\xi}\sqrt{\rho-1}\norm{\bar{\vg}_t}.
\end{align*}
Therefore,
\begin{align*}
    \bar{\vg}_t^\top\E_t[\widetilde{\vg}_t]
    &=
    \bar{\vg}_t^\top\E_t[\xi_t\vg_t]
    \\
    &\ge
    \xi_{\min}\norm{\bar{\vg}_t}^2
    -
    \bar{\sigma}_{\xi}\sqrt{\rho-1}\norm{\bar{\vg}_t}^2
    \\
    &=
    \mu\norm{\bar{\vg}_t}^2.
\end{align*}
This proves the alignment bound. For the second moment,
\begin{align*}
    \E_t\big[\norm{\widetilde{\vg}_t}^2\big]
    &=
    \E_t\big[\xi_t^2\norm{\vg_t}^2\big]
    \\
    &\le
    \xi_{\max}^2\E_t\big[\norm{\vg_t}^2\big]
    \\
    &\le
    \xi_{\max}^2\rho\norm{\bar{\vg}_t}^2
    =
    M_2\norm{\bar{\vg}_t}^2.
\end{align*}
\end{proof}

\begin{remark}[Conservativeness of the sufficient condition]
The positivity condition $\mu>0$ can be written as
\begin{align*}
    \xi_{\min}
    >
    \frac{1}{2}(\xi_{\max}-\xi_{\min})\sqrt{\rho-1}.
\end{align*}
Equivalently, with
$A=(\Lambda_{ff}+\epsilon_c)Q$, a sufficient condition is
\[
    \epsilon_{\mathrm{LM}}
    >
    \frac{A}{2}\sqrt{\rho-1}.
\]
This is a sufficient worst-case covariance condition.
The constant $\mu$ measures the worst-case alignment retained by the expected
scaled stochastic gradient, while $M_2$ controls its conditional second moment.
The condition can be conservative, especially when the bound
$\xi_t\le1/\epsilon_{\mathrm{LM}}$ is loose.
The main result therefore states Assumption~\ref{ass:alignment-moment}
directly and uses this proposition only as one way to verify it.
\end{remark}

\section{Experiment details}\label{apx:exp_details}

\subsection*{Updates used in the experiments}

Algorithm~\ref{alg:ignd} gives the generic supervised IGND update. 
For a squared Bellman error with target $y_t$ treated as fixed during differentiation, 
define
\[
    \delta_t = y_t-q_{\vw_t}(\vs_t,a_t),
    \qquad
    \vj_t=\nabla_{\vw}q_{\vw_t}(\vs_t,a_t).
\]
The corresponding semi-gradient IGND update is
\[
    \vw_{t+1}
    =
    \vw_t
    +
    \alpha_t
    \frac{\delta_t}{\norm{\vj_t}^2+\epsilon_{\mathrm{LM}}}
    \vj_t.
\]
When $y_t$ is bootstrapped, this is a semi-gradient update rather than the
exact gradient step for a fixed empirical risk objective.

For Adam-IGND, let
\[
    \vg_t=\ell_{f,t}\vj_t,
    \qquad
    \xi_t=
    \frac{1}
    {(\ell_{ff,t}+\epsilon_c)\norm{\vj_t}^2+\epsilon_{\mathrm{LM}}}.
\]
The direction passed to Adam is the IGND-scaled gradient
\[
    \widetilde{\vg}_t=\xi_t\vg_t.
\]
Adam-IGND then uses the same update equations as the Adam baseline:
\begin{align*}
    \vm_t
    &=
    \beta_1\vm_{t-1}+(1-\beta_1)\widetilde{\vg}_t,\\
    \vv_t
    &=
    \beta_2\vv_{t-1}
    +(1-\beta_2)\widetilde{\vg}_t\odot\widetilde{\vg}_t,\\
    \widehat{\vm}_t
    &=
    \frac{\vm_t}{1-\beta_1^t},
    \qquad
    \widehat{\vv}_t
    =
    \frac{\vv_t}{1-\beta_2^t},\\
    \vw_{t+1}
    &=
    \vw_t
    -
    \alpha_t
    \frac{\widehat{\vm}_t}
    {\sqrt{\widehat{\vv}_t}+\epsilon_{\mathrm{Adam}}}.
\end{align*}
Adam and Adam-IGND use the same fixed Adam constants
$\beta_1=0.9$, $\beta_2=0.999$, and
$\epsilon_{\mathrm{Adam}}=10^{-8}$; 
only the learning rate is selected by validation performance.

\subsection*{Controlled scale robustness test}

The FrozenLake experiment uses \texttt{FrozenLake-v1}~\cite{openai_gym} 
with deterministic transitions and a tabular one-hot representation 
for the state-action value function. 
In the scaled feature condition, each coordinate is multiplied by a
fixed random nonzero integer sampled uniformly from $[-10^4,10^4]$. 
The same scaling is used for all methods and seeds. 
This keeps the tabular task fixed
while changing the local output sensitivity of the parameterization.

The main comparison uses QL and IGNDQ with the same learning rate $\alpha=1$. 
In the scaled feature condition, an additional QL curve with
$\alpha=10^{-8}$ is included as a reference.

\subsection*{Supervised learning}

California Housing and Diamonds are regression tasks trained with squared
loss. 
Cats vs Dogs and IMDB Reviews are binary classification tasks trained
with binary cross-entropy with logits. 
Numerical features are scaled,
categorical features are one-hot encoded, 
images are resized to
$224\times224$ and scaled to $[0,1]$, 
and text is tokenized using the \texttt{cl100k\_base} byte-pair encoding.

For regression, the model is a feedforward network with hidden widths $32$,
$64$, and $32$ and ReLU activations. 
For Cats vs Dogs, the model is a CNN with
three convolutional layers with $32$, $64$, and $128$ channels, $3\times3$
kernels, ReLU activations, and $2\times2$ max pooling, followed by fully
connected layers of width $128$ and $64$. 
For IMDB Reviews, the model uses an
embedding layer of dimension $128$ followed 
by an LSTM~\cite{hochreiter1997long} with $128$ units.

The supervised experiments compare SGD, IGND, Adam, and Adam-IGND.
Adam-IGND first applies the IGND scalar normalization to the sample gradient
and then passes the scaled direction to Adam, as described above. 
Adam-IGND is included only as an empirical compatibility test 
and is not part of the theoretical development.

The regression metric is mean absolute percentage error,
\[
    \mathrm{MAPE}
    =
    \frac{1}{N}\sum_{i=1}^{N}
    \frac{\abs{y_i-f_{\vw}(\vx_i)}}{\max\{10^{-15},\abs{y_i}\}}.
\]
Binary classification is evaluated by accuracy.

\subsection*{Linear-quadratic control}

The LQR experiments use two systems from the Compleib benchmark set
\cite{leibfritz2006compleib}: the linearized vertical-plane dynamics of an
aircraft (UAV)~\cite{hung1982multivariable} and a binary distillation tower
(BDT)~\cite{davison1990benchmark}. 
Both systems are discretized with sampling period
$\Delta T=0.1\,\mathrm{s}$. For state dimension $n_s$ and action dimension
$n_a$, the reward matrices are $\mQ=-\mI_{n_s}$ and $\mR=-\mI_{n_a}$.

We follow the policy iteration approach for learning LQR from data described
in~\cite{bradtke1994adaptive,bradtke1996linear}. 
In this approach, policy evaluation estimates a quadratic action-value function, 
and policy improvement forms the greedy linear feedback policy.

For policy evaluation, 
the action value is represented as a linear function of
quadratic features of the state-action pair. 
Let
$\vz_t=[\vs_t^\top,\va_t^\top]^\top$ and let $\vx_t=\vx(\vs_t,\va_t)$ 
collect the upper-triangular quadratic features of $\vz_t\vz_t^\top$. 
In stochastic variants, a constant bias feature is appended. 
Thus
\[
    q_{\vw}(\vs_t,\va_t)=\vw^\top\vx_t.
\]
Equivalently, after converting the learned weights to a symmetric matrix,
the quadratic part can be written as
\[
    q_{\mM}(\vs_t,\va_t)=\vz_t^\top\mM\vz_t
\]
up to the constant bias term used in the stochastic variants.

For a fixed policy $\mK_i$, the next action is
$\va'_t=\mK_i\vs_{t+1}$, and the Bellman target is treated as fixed during the
semi-gradient update. With
\[
    \vx'_t=\vx(\vs_{t+1},\va'_t),
    \qquad
    \delta_t
    =
    r_t+\gamma\vw_t^\top\vx'_t-\vw_t^\top\vx_t,
\]
the QL policy evaluation update is
\[
    \vw_{t+1}
    =
    \vw_t+\alpha_t\delta_t\vx_t,
\]
whereas the IGNDQ policy evaluation update used in these experiments is
\[
    \vw_{t+1}
    =
    \vw_t+
    \alpha_t
    \frac{\delta_t}{\norm{\vx_t}^2}
    \vx_t.
\]
This is the undamped squared Bellman error normalization.

After policy evaluation, the learned weights are converted to the symmetric
matrix $\mM$ and partitioned as
\[
    \mM
    =
    \begin{bmatrix}
        \mM_{ss} & \mM_{sa}\\
        \mM_{as} & \mM_{aa}
    \end{bmatrix}.
\]
When $\mM_{aa}\prec 0$, the greedy linear feedback policy is
\[
    \mK_{i+1}
    =
    -\mM_{aa}^{-1}\mM_{as}.
\]
This negative definiteness condition is the finite maximizer condition under
the reward convention $\mR=-\mI_{n_a}$.

The plotted quantity is the Frobenius policy error
$\norm{\mK^\star-\mK_i}_{F}$, 
where $\mK^\star$ is the optimal LQR
policy and $\mK_i$ is the policy after the $i$th recorded policy improvement step. 
Curves are plotted over the recorded policy improvement iterations for
each method; 
no padding or extrapolation is applied when a run records fewer iterations.

\end{document}